\documentclass[letterpaper]{article} 
\usepackage{aaai24}  
\usepackage{times}  
\usepackage{helvet}  
\usepackage{courier}  
\usepackage[hyphens]{url}  
\usepackage{graphicx} 
\usepackage{natbib}  
\usepackage{caption} 
\usepackage{algorithm}
\usepackage{algorithmic}
\usepackage[utf8]{inputenc} 
\usepackage[T1]{fontenc}    
\usepackage{url}            
\usepackage{booktabs}       
\usepackage{amsfonts}       
\usepackage{nicefrac}       
\usepackage{microtype}      
\usepackage{xcolor}         
\usepackage{xspace}
\usepackage{amsmath}
\usepackage{multirow}
\usepackage{rotating}
\usepackage{tabularx}
\usepackage{gensymb}
\usepackage{amssymb}
\usepackage{graphicx}
\usepackage{float}
\usepackage{floatflt}
\usepackage{makecell}
\usepackage{colortbl}
\usepackage{calc}

\definecolor{myPurple}{rgb}{0.4, .0, .8}
\definecolor{myGreen}{rgb}{0, .8, .3}
\definecolor{myRed}{rgb}{0.8, .2, .2}
\definecolor{myBlue}{rgb}{0.0, .0, .8}
\definecolor{springgreen}{RGB}{204, 255, 51}
\definecolor{myWhite}{RGB}{255, 255, 255}

\newcommand{\boldparagraph}[1]{\noindent{\bf #1} }

\colorlet{colorFst}{myGreen!25}       
\colorlet{colorSnd}{springgreen!45} 
\colorlet{colorblank}{myWhite}  
\newcommand{\red}[1]{ \noindent {\color{myRed}{} {#1}}}

\newcommand{\fs}[1]{\colorbox{colorFst}{\textbf{#1}}}
\newcommand{\nd}[1]{\colorbox{colorSnd}{\textbf{#1}}}     
\newcommand{\bl}[1]{\colorbox{colorblank}{#1}}

\definecolor{myred}{rgb}{1.0,0.0,0.0}

\makeatletter
\DeclareRobustCommand\onedot{\futurelet\@let@token\@onedot}
\def\@onedot{\ifx\@let@token.\else.\null\fi\xspace}

\def\eg{\emph{e.g}\onedot} 
\def\ie{\emph{i.e}\onedot} 
 
\def\etc{\emph{etc}\onedot} 
 
\def\etal{\emph{et al}\onedot}

\usepackage{newfloat}
\usepackage{listings}
\DeclareCaptionStyle{ruled}{labelfont=normalfont,labelsep=colon,strut=off} 
\floatstyle{ruled}
\newfloat{listing}{tb}{lst}{}
\floatname{listing}{Listing}
\title{PNeRFLoc: Visual Localization with Point-based Neural Radiance Fields}
\author{
    Boming Zhao\textsuperscript{\rm 1},
    Luwei Yang,
    Mao Mao\textsuperscript{\rm 1},
    Hujun Bao\textsuperscript{\rm 1},
    Zhaopeng Cui\textsuperscript{\rm 1}\thanks{Corresponding author.} 
}
\affiliations{
    \textsuperscript{\rm 1}State Key Lab of CAD \& CG, Zhejiang University\\
}

\usepackage{bibentry}

\begin{document}

\maketitle
\begin{abstract}
Due to the ability to synthesize high-quality novel views, Neural Radiance Fields (NeRF) has been recently exploited to improve visual localization in a known environment. However, the existing methods mostly utilize NeRF for data augmentation to improve the regression model training, and their performances on novel viewpoints and appearances are still limited due to the lack of geometric constraints. In this paper, we propose a novel visual localization framework, \ie, PNeRFLoc, based on a unified point-based representation. On one hand, PNeRFLoc supports the initial pose estimation by matching 2D and 3D feature points as traditional structure-based methods; on the other hand, it also enables pose refinement with novel view synthesis using rendering-based optimization. Specifically, we propose a novel feature adaption module to close the gaps between the features for visual localization and neural rendering.
To improve the efficacy and efficiency of neural rendering-based optimization, we also developed an efficient rendering-based framework with a warping loss function. 
Extensive experiments demonstrate that PNeRFLoc performs the best on the synthetic dataset when the 3D NeRF model can be well learned, and significantly outperforms all the NeRF-boosted localization methods with on-par SOTA performance on the real-world benchmark localization datasets. 
The code and supplementary material are available on the project webpage: \red{https://zju3dv.github.io/PNeRFLoc/}. 
\end{abstract}

\section{Introduction}
\label{sec:introduction}

Visual localization is a fundamental task in computer vision that aims to determine the precise position and orientation of a camera in a known scene based on the visual input, and it has widespread applications in areas such as robot navigation, augmented reality, virtual reality, \etc.
Traditional structure-based localization, as the mainstream solution for visual localization, has advantages such as scene agnosticism, robustness, and high precision. These methods ~\cite{dsac++, hloc} require computing and storing a global map consisting of 3D point
locations and try to find the correspondences between 2D feature points extracted in the query image and 3D points in the reconstructed scene and use a Perspective-n-Point (PnP) solver 
\cite{p3p_review, p4p} 
in a RANSAC loop \cite{ransac, random_ransac}  
to compute the camera poses. Besides hand-crafted features \cite{bay2008speeded, lowe2004distinctive}, deep features ~\cite{detone2018superpoint, dusmanu2019d2, s2dnet, superglue} 
have been extensively utilized recently to improve feature matching for better localization. Very recently, some state-of-the-art (SOTA) feature matching methods have been proposed to train the deep features and align features through pose refinement in an end-to-end manner~\cite{pixel-perfect, sarlin2021back}. 
However, these structure-based methods rely on 2D-3D or 2D-2D point matching, and thus the accuracy is limited when the feature extraction and matching is sparse or noisy due to the large view changes between images and textureless structures. 

Regression-based localization trains a neural network and takes the network parameters as a global map representation which can directly regress the 6-DOF camera poses
~\cite{kendall2015posenet, relocnet, posenet_pgo, moreau2022coordinet, TransPoseNet} 
or the 3D scene coordinate of each pixel 
\cite{cavallari2017fly, li2020hierarchical, sanet} 
by taking the query image as the network input.
For the simplicity and end-to-end training manner, these methods have attracted considerable attention. However, these methods are usually scene-specific, and the accuracy heavily relies on the distribution of the training images with poor generalization to new viewpoints~\cite{sarlin2021back}. 
Thus Neural Radiance Fields (NeRF)~\cite{mildenhall2020nerf} has been introduced recently to render realistic novel viewpoint images for data augmentation ~\cite{chen2022dfnet, direct_posenet, moreau2022lens} that can boost the training of the regression network. However, these methods are inherently regression-based, which imposes constraints on the localization accuracy as it is not feasible to indefinitely expand the training data and cover the whole 6D pose space. 

To fix the problems of existing methods, we propose a novel framework for visual localization, \ie, PNeRFLoc, that integrates the structure-based framework and the rendering-based optimization with NeRF representation. Specifically, on one hand, our framework supports the initial pose estimation by matching 2D and 3D feature points; on the other hand, it also enables the pose refinement with novel view synthesis using rendering-based optimization, \ie, minimizing the photometric error between the rendered image and the query image. In this way, compared to the existing NeRF-boosted methods \cite{chen2022dfnet, direct_posenet, moreau2022lens}, our approach transcends the limitations of regression-based techniques, achieving significant accuracy improvements in both indoor and outdoor scenes. Moreover, compared to the SOTA feature matching methods which may be limited by sparse matches between reference and query images due to large view changes and thus stuck in local optima, our method can achieve better accuracy by minimizing the photometric loss with the capability to render novel-view images.

However, it is non-trivial to design the framework. First, there is no unified scene representation that supports both 2D-3D feature matching used in structure-based localization and neural rendering for rendering-based optimization.
In this paper, we adapt a recent point-based neural radiance field representation (\ie, PointNeRF \cite{xu2022point}) and design a feature adaptation module to bridge the gap between the scene-agnostic features for localization and the scene-specific features for neural rendering. We find that although these two types of features aim for different tasks, they can be easily transferred via a feature adaptation module. In this way, we can utilize any existing scene-agnostic features for initial localization (\eg, R2D2 \cite{r2d2}), and learn the scene-specific adaptation module together with the NeRF models. 
Moreover, from the adaptation module, we can also learn a score for each dense feature for better feature matching and initial localization. 
Second, the rendering-based optimization may be easily stuck in the local minimum \cite{maggio2022loc} due to the backpropagation through the networks and also time-consuming. To improve the neural rendering-based optimization with point-based representation, we further propose a novel efficient rendering-based optimization framework by aligning the rendered image with the query image and minimizing the warping loss function. 
In this way, we don't need to render a new image for each step of optimization and avoid the backpropagation through the networks for better convergence. 
Lastly, to further improve the robustness of the proposed method for outdoor illumination changes and dynamic objects, we utilize appearance embedding and segmentation masks to handle varying lighting conditions and complex occlusions respectively.

Our contributions can be summarized as follows. At first, we propose a novel visual localization framework with a unified scene representation, \ie, PNeRFLoc, which enables both structure-based estimation and render-based optimization for robust and accurate pose estimation. 
Second, to close the gaps between the features for visual localization and neural rendering, we propose a novel feature adaptation module that can be learned together with NeRF models. Furthermore, a novel efficient rendering-based framework with a warping loss function is proposed to improve the efficacy and efficiency of neural rendering-based optimization. 
Extensive experiments show that the proposed framework outperforms existing learning-based methods when the NeRF model can be well learned, and performs on-par with the SOTA method on the visual localization benchmark dataset. 

\section{Related Work}
\label{sec:related_work}
\boldparagraph{Structure-based localization.} Structure-based methods~\cite{camposeco2017toroidal,Cheng_2019_ICCV,sattler2015hyperpoints,sattler2016efficient,Toft_2018_ECCV,Zeisl_2015_ICCV} utilize 3D scene information from structure from motion (SfM), and a query image taken from the same scene can be registered with explicit 2D-3D correspondences and PnP + RANSAC algorithm.
Typically, these methods can yield accurate poses but are prone to noisy matches. To mitigate outlier influence, recent scene coordinates regression \cite{dsac, dsac++, sanet} methods rely on CNNs to fuse semantic features for obtaining accurate dense correspondences map, while the recent ~\cite{superglue} excels graphical transformer with 2D relative positional encoding to achieve impressive sparse matching results. Despite the promising performance, the structure-based methods still suffer large-view changes especially when a few reference points are available.

\boldparagraph{Regression-based localization.} PoseNet~\cite{kendall2015posenet} and its subsequent work~\cite{walch2017image} regress the camera pose of an image directly through CNN or LSTM. These methods are limited in terms of scalability and performance. Despite some attempts to improve the accuracy by incorporating geometry prior~\cite{MapNet}, these methods can only perform comparable results to that of image retrieval baselines~\cite{NetVLAD, DenseVLAD} and cannot achieve the identical performance of structure-based counterparts. Moreover, adapting these regressed models to novel scenes is prohibited, which narrows their potential for real-time applications.

\boldparagraph{Localization with NeRF.}  
Neural Radiance Fields~\cite{mildenhall2020nerf} has recently been employed for localization tasks. This is because NeRF can synthesize high-quality novel view images, which can be beneficial for localization tasks. For example, Purkait \etal proposed LENS~\cite{moreau2022lens}, which uses NeRF-w~\cite{martin2021nerf} to render realistic synthetic images to expand the training space. LENS leverages the NeRF-w model to obtain scene geometry information and render views from virtual camera poses covering the entire scene. However, LENS is limited by its long-time offline pre-training and infeasibility of covering the whole pose space, 
and it also lacks compensation for the domain gap between synthetic and real images, such as pedestrians and vehicles in outdoor scenes. 
Chen \etal proposed DFNet~\cite{chen2022dfnet}, which incorporates an additional feature extractor to learn high-level features to bridge the domain gap between synthetic and real images. However, the training process remains lengthy, as DFNet still needs to train NeRF, pose regression, and feature extraction networks separately. Maggio \etal proposed a Monte Carlo localization method called Loc-NeRF~\cite{maggio2022loc}, where Loc-NeRF continuously samples candidate poses under the initial pose and uses NeRF to render novel views to find the correct pose direction. However, Loc-NeRF is unstable and still requires an initial camera pose. Moreover, Yen-Chen \etal introduced iNeRF~\cite{yen2021inerf}, an inverse NeRF approach to optimize camera poses, but it is also limited by the need to provide an initial pose.

\begin{figure*}[h]
  \centering
  \includegraphics[width=0.98\linewidth, trim={0 0 0 0}, clip]{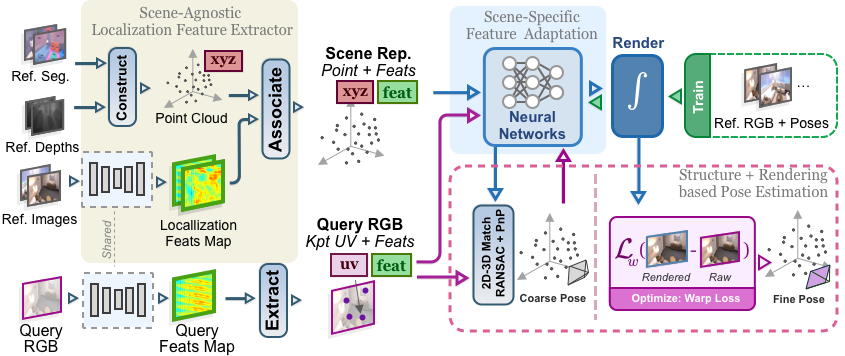}
  \vspace{-1.0em}
  \caption{\textbf{Visual localization with PNeRFLoc}. In the proposed framework, we associate raw point clouds with scene-agnostic localization features and train a scene-specific feature adaptation together with the point-based neural radiance fields. 
  Subsequently, PNeRFLoc integrates structure-based localization with novel rendering-based optimization to accurately estimate the 6-DOF camera pose of the query image.}
  \label{fig:pipeline}
  \vspace{-1.5em}
\end{figure*}

\section{Method}

We propose a novel visual localization framework called \textbf{PNeRFLoc} based on the scene representation as shown in Fig.\ref{fig:pipeline}. In order to enable both structure-based estimation and render-based optimization in a unified framework, we adapt the recent point-based radiance field representation~\cite{xu2022point} and design a feature adaptation module to bridge the scene-agnostic localization feature and the point-based neural rendering (Sec. 3.2). Additionally, to prevent iterative re-rendering of images for every optimization step like iNeRF~\cite{yen2021inerf}, we propose an efficient rendering-based optimization strategy by minimizing the warping loss function to align the pixels on the rendered image and the query image, 
which reduced the neural rendering frequency to just once for most cases,
while performing high accuracy~(Sec. 3.3). 
\label{ssec:baseline}
\subsection{Point-based Radiance Field Representation} 


Neural Radiance Fields (NeRF) compute pixel radiance by sampling points along the ray shot through each pixel and computing the integral result. Specifically, each pixel in an image corresponds to a ray $\mathbf{r}(t) = \mathbf{o} + t\mathbf{d}$.
To render the color of ray $\mathbf{r}$, NeRF draws the point samples with distances $\{t_i\}^N_{i=1}$ to the camera origin $\mathbf{o}$ along the ray, and passes the point locations $\mathbf{r}(t_i)$ as well as view directions $\mathbf{d}$ to obtain density $\sigma_i$ and colors $\mathbf{c}_i$.
The resulting color is rendered following the quadrature rules~\cite{max1995optical}: 
\vspace{-0.8em}
\begin{footnotesize}
\begin{equation}
\label{eq:render}
\begin{split}
    \hat{C}(\mathbf{r}) = \mathcal{R}(\mathbf{r},\mathbf{c},\sigma) = 
    \sum^K_{k=1}T(t_k)\alpha\left(\sigma(t_k)\delta(t_k)\right)\mathbf{c}(t_k), \\
    T(t_k)={\rm exp}(-\sum_{k'=1}^{k-1}\sigma(t_{k'})\delta_{k'}), \;\; 
    \alpha(x) = 1 - {\rm exp}(-x),
\end{split}
\end{equation}
\end{footnotesize}
where $\mathcal{R}(\mathbf{r},\mathbf{c},\sigma)$ is the volumetric rendering through ray $\mathbf{r}$ of color $\mathbf{c}$ with density $\sigma$, $\mathbf{c}(t)$ and $\sigma(t)$ are the color and density at point $\mathbf{r}(t)$ respectively, and $\delta_k = t_{k+1} - t_k$ is the distance between two adjacent sampling points on the ray. Stratified sampling and informed sampling are used to select sample points $\{t_k\}^K_{k=1}$ between the near plane $t_n$ and far plane $t_f$. Additionally, 
the depth $\hat{D}$ of each ray $\mathbf{r}$ can be computed as: 
\vspace{-0.5em}
\begin{equation}
\label{eq:render_depth}
\footnotesize
    \hat{D} = \sum^K_{k=1}T(t_k)\alpha\left(\sigma(t_k)\delta(t_k)\right)t_k.
\end{equation}
Following PointNeRF~\cite{xu2022point}, we regress the radiance field from the point cloud $P = \{(p_i, f_i, \gamma_i)|i=1,...,N\}$, where each point $i$ is located at $p_i$ and associated with a feature vector $f_i$ that encodes the local scene content. And $\gamma_i$ represents the confidence of a point being located on the actual surface of the scene. Given any 3D location $x$, we query $K$ neighboring neural points around $x$ and regress the density $\sigma$ and view-dependent color $c$ from any viewing direction $d$ as:
\begin{equation}
\footnotesize
(\sigma, c) = \mathbf{PointNeRF}(x,d,p_1,f_1,\gamma_1,...,p_K,f_K,\gamma_K).
\end{equation}

To enable PointNeRF to handle dynamic objects and illumination changes, we adopt the appearance embedding from NeRF-W~\cite{martin2021nerf} and a segmentation mask to handle occasional object occlusions and illumination variations. Contrary to NeRF-W which directly employs a transient MLP to address the issue of occasional object occlusions, we adopt a more stable approach by utilizing a segmentation mask to compel the network to focus exclusively on architectural areas. In this paper, we utilize Detectron2\footnote{\url{https://github.com/facebookresearch/detectron2}} to perform object detection on the images and generate segmentation masks. 

\vspace{-0.2em}
\subsection{Scene-Specific Feature Adaptation}
\label{ssec:transfer}

Once the point-based NeRF model is built, a straightforward way is to utilize the learned point-wise neural features for feature matching. However, we find that these neural features are not distinctive enough and cannot be used as robust and efficient descriptors for feature matching as shown in our supplementary material because these features are learned to encode local color and geometry information of the specific scene for the neural rendering.
Based on this observation, we resort to existing well-studied deep features for visual localization that are trained on large datasets for feature matching and design a feature adaptation module to bridge the features for visual localization and neural rendering. Moreover, we can also learn the scores of neural features with the adaptation module for better feature matching for the specific scene.

\boldparagraph{Scene-agnostic point localization feature extractor.}
In this paper, we utilize the deep feature R2D2 \cite{r2d2} as the scene-agnostic feature for visual localization due to its ability to robustly extract reliable and distinctive features. For each reference image $I_k\in\mathbb{R}^{W \times H \times 3}$, the R2D2 network extracts a feature map $\mathbf{F}_k\in \mathbb{R}^{W \times H \times 128}$. For each 3D point $i$ constructed from each reference image $k$, we define the scene-agnostic point feature as: 
\begin{equation}
\footnotesize
    f_i = \mathbf{F}_k[p_i] \in \mathbb{R}^{128},
\end{equation}
where $p_i$ is the projection of $i$ in the reference image and $[\cdot]$ is a lookup with sub-pixel interpolation. Searching for matches throughout the entire point cloud is inefficient 
as the reliability of scene-agnostic localization features can be compromised by the structure of the scene. 
Therefore, we utilize the matching score of each point (introduced in the feature adaptation) to enable point filtering during the matching process.
By removing candidate points in the point cloud below a certain score threshold, we can reduce the number of matching pairs to be computed, thus improving efficiency. 

\boldparagraph{Scene-specific feature adaptation.}
As explained before, due to the significant gap between scene-agnostic features for localization and scene-specific features for neural rendering, we cannot learn the radiance fields from the R2D2 features. 
Thus we design a feature adaptation module to bridge this gap, which consists of a four-layer Multi-Layer Perceptron (MLP).
We empirically find that despite the scene-agnostic feature and the scene-specific NeRF representation feature aiming at two completely different tasks, they can be adapted via the designed module. Thus any other SOTA scene-agnostic feature for visual localization can also be utilized in our framework.
Moreover, as mentioned above, we also utilize the adaptation module to learn a score $S$ for each point in the point cloud according to its dense features and position. These scores are then used for point filtering, which improves the efficiency of feature matching while maintaining the accuracy of the final pose estimation.

\boldparagraph{Scene-specific PointNeRF reconstruction.}
Given the pre-trained point localization feature extractor and feature adaptation module, similar to PointNeRF \cite{xu2022point}, we learn the NeRF model by minimizing the following loss function:
\begin{equation}
\footnotesize
    \mathcal{L}_{render} = \sum_{\textbf{r}\in R} \lVert\hat{C}(\textbf{r}) - C(\textbf{r}) \rVert^2_2,
\end{equation}
where $R$ is the set of rays in each batch, and $C(r)$, $Cˆ(r)$ are the ground truth and predicted RGB colors for ray $r$ computed by Eq.\ref{eq:render}. To be noted, we also learned to fine-tune the feature adaptation module for each scene for better rendering quality. Please refer to our supp. material for more details. 
\vspace{-0.5em}

\subsection{Two-stage Pose Estimation} \label{ssec:Loc-details}

Once the NeRF model is learned, we design a two-stage pose estimation framework for the query image during the test. 

\boldparagraph{Initialization with structure-based localization.}
The goal of the structure-based localization stage is to establish the correspondence between the 2D key points on the query image and the 3D points in the scene point cloud, thereby providing an initial pose estimate for the subsequent pose refinement stage. For each query image $q$ with the keypoints $P_q$ and features $\mathbf{F}_q$ extracted by the scene-agnostic localization feature extractor, and the point cloud $P_r$ generated by PointNeRF with features $\mathbf{F}_r$, we can find a 2d-3D correspondence:
\begin{equation}
\footnotesize
\forall i \in P_q, \quad M(i) = \arg\max_{j \in P_r} \frac{\mathbf{F}^{i}_{q} \cdot \mathbf{F}^{j}_{r}}{\|\mathbf{F}^{i}_{q}\| \|\mathbf{F}^{j}_{r}\|},
\end{equation}
where $M(i)$ signifies the corresponding point within the point cloud for the keypoint $i$ present on the query image q, which is ascertained via the maximization of cosine similarity. However, as mentioned in Sec. 3.2, the process of directly seeking correspondences within the entire point cloud proves to be inefficient. In response to this, we employ a thresholding technique based on 
the learned score $S$ to filter the point cloud $P_r$. Given a threshold $S_t$, we can get the filtered point cloud as $P_s = \{i \in P_r \mid S_i \geq S_t\}$, where $S_i$ denotes the learned score for 
the point $i$.
For each query image q and the discovered correspondence $M$, we define a residual:
\begin{equation}
\footnotesize
r_i = \lVert p_i - \prod{(\mathbf{R}M(i) + \mathbf{t})} \rVert_2,
\end{equation}
where $\prod({\cdot})$ represents the pixel obtained post the projection of the 3D point onto the image. $(\mathbf{R}, \mathbf{t})$ denote the camera pose to be determined, while $p_i$ signifies the pixel of the keypoint $i$ within the query image. The total error over all key points is: 
\begin{equation}
\label{eq:pnp}
\footnotesize
E(\mathbf{R}, \mathbf{t}) = \sum_{i \in P_q}{r_i}.
\end{equation}
Moreover, a direct optimization of Eq.\ref{eq:pnp} is often susceptible to the distortions caused by incorrect correspondences (outliers). Therefore, a RANSAC loop is also adopted, effectively improving the accuracy.

\boldparagraph{Pose refinement with efficient rendering-based optimization.}
Previous works~\cite{yen2021inerf, zhu2022nice} have utilized gradient descent to minimize the photometric residuals between the rendered and input images for local pose estimation. However, this optimization method is inefficient since neural rendering is required for each optimization step, and it is also unstable due to the backpropagation over the deep networks. 
Therefore, we propose a novel and efficient rendering-based optimization strategy using the warping loss function, which only requires rendering the image once and avoiding the backpropagation through the networks. 

Specifically, for a given query image $q$ and initial pose $(\mathbf{R},\mathbf{t})$, PNeRFLoc first renders the visual reference image $q_r$ under the initial pose according to Eq.~\eqref{eq:render} and the depth map $d_q$ according to Eq.~\eqref{eq:render_depth}. Subsequently, we randomly sample $N$ pixels within the image $q_r$. For the pose $(\mathbf{R'}, \mathbf{t'})$ that we aspire to optimize, we define the warping loss function as:
\begin{equation}
    \footnotesize
    \mathcal{L}_{warping} = \sum_{p_i\in N} \lVert C(q, W(p_i, \mathbf{R}, \mathbf{t}, \mathbf{R'}, \mathbf{t'})) - C(q_r, p_i) \rVert_2, 
\end{equation}
\vspace{-0.5em}
\begin{equation}
    \footnotesize
    W(p_i, \mathbf{R}, \mathbf{t}, \mathbf{R}', \mathbf{t}') = \prod(\mathbf{R}'(\mathbf{R}^{-1}\prod\nolimits^{-1}(p_i, \hat{D}(p_i) - \mathbf{R}^{-1}\mathbf{t}) + \mathbf{t}'),
\end{equation}
where $C(q_r, p_i)$ represents the RGB color at pixel $p_i$ on rendering image $q_r$, and the function $W$ denotes the corresponding pixel on query image $q$ by warping $p_i$ from render image $q_r$. Specifically, $W$ back-projects $p_i$ into the 3D space of the $q_r$'s camera coordinate system using the depth $\hat{D}(p_i)$, and then transposes it to the camera coordinate system of image $q$ through the camera pose $(\mathbf{R'}, \mathbf{t'})$ and projects it onto image $q$ finally. However, we find that there are often blanks in the rendering images, which occur when the rays emitted from the camera pass through the gaps in the point cloud and do not aggregate to the neural points. In this case, incorrect depth and color can interfere with the optimization. Hence, we propose using a blank depth mask to handle such situations. For the set of sampled pixels $N$ on the visual reference image $q_r$, we define the valid pixels set as $N_v = \{p_i \in N | \hat{D}(p_i) >= 0.01\}$ and let the warping loss function only consider the pixels in $N_v$. Our rendering-based optimization method optimizes the pose $(\mathbf{R'}, \mathbf{t'})$ by using warping loss aligning the RGB colors of sampled pixels on $q_r$ and $q$. Thereby we avoid gradient descent through the complex neural networks and improve accuracy and efficiency. Moreover, when the viewpoint changes significantly between the visual reference image and the query image, the optimization result may not reach its optimum. We could potentially enhance the accuracy by iteratively rendering the visual reference image multiple times. However, we found that a single rendering's outcome was already satisfactory in our experiments. 
Therefore, to save time, our optimization process only renders the visual reference image once in practice.

\vspace{-0.5em}
\section{Experiments}

We first compare our method with various representative and SOTA learning approaches \cite{sarlin2021back, moreau2022coordinet, moreau2022lens, dsac++} on both synthetic datasets and real-world datasets. Then, we offer insights into PNeRFLoc through additional ablation experiments.

\vspace{-0.5em}
\subsection{Datasets and Implementation Details}
\paragraph{Datasets.} Following \cite{chen2022dfnet, moreau2022lens, moreau2022coordinet}, we evaluate our method on two standard localization datasets since they have well-distributed training images which support dense 3D reconstruction. Moreover, we generate a synthetic localization dataset using the commonly used Replica dataset in NeRF-based SLAM systems.
\begin{itemize}
\item \textbf{Cambridge Landmarks~\cite{kendall2015posenet}} contains five outdoor scenes, with 200 to 2000 images captured at different times for each scene. This dataset is challenging for camera pose estimation because the query images are taken at different times than the reference images, resulting in different lighting conditions and occlusions from objects such as people and vehicles. 
\item \textbf{7Scenes~\cite{shotton2013scene}} contains seven indoor scenes, captured by a Kinect RGB-D sensor. Each scene has 1k to 7k reference images and 1k to 5k query images, captured along different trajectories. 
\item \textbf{Replica~\cite{straub2019replica}} contains eight synthetic indoor scenes, commonly used for SLAM evaluation. We follow iMAP \cite{sucar2021imap}, using its produced sequences as training images, with an image size of 1200*680 pixels, and randomly generate 50-120 query images. Due to the small number of reference images and the significant changes in the viewpoint of the query images, this dataset presents a certain level of challenge for localization tasks.
\end{itemize}

\boldparagraph{Implementation.} 
%
We use R2D2~\cite{r2d2} as the scene-agnostic localization feature extractor. In the structure-based localization stage, the score threshold $S_t$ is set to 0.7, and the number of RANSAC iterations is set to 20k. During the rendering-based localization stage, we use the Adam optimizer with a learning rate of 0.001. All our experiments are evaluated on a single NVIDIA GeForce RTX 3090 GPU. 
Similar to DSAC*~\cite{dsac++}, we obtain the estimated depth images rendered from a 3D model learned by Factorized-NeRF~\cite{zhao2022factorized}.  For the 7scenes dataset, we follow PixLoc~\cite{sarlin2021back} utilizing the estimated depth rendered by DSAC*~\cite{dsac++}. Lastly, we leveraged a pre-trained model from MonoSDF~\cite{yu2022monosdf} to render the estimated depth for the Replica dataset.
Please refer to the supp. material for more 
details.
\vspace{-0.7em}
\subsection{Evaluation on the Replica Dataset}
\vspace{-0.2em}
\label{ssec:exp_sota}
\begin{table}[!t]
  \caption{\textbf{Comparison on Replica datasets. 
  } We report median translation/rotation errors~(meters/degrees) and the best results are highlighted as \fs{\bf{first}}.}
  \label{SOTA table}
  \centering
  \small
  \vspace{-1.0em}
  \begin{tabular}{clccc}
    \toprule
    \multicolumn{1}{c}{} & Methods & CoordiNet & PixLoc &\textbf{PNeRFLoc}\\
    \midrule
    \multirow{8}{*}{\rotatebox{90}{Replica}} & room0 & \bl{1.60}/\bl{50.8} & \bl{0.055}/\bl{1.89} & \fs{\bf{0.005}}/\fs{\bf{0.29}}\\
    & room1 & \bl{1.38}/\bl{47.3} & \bl{0.020}/\fs{\bf{0.36}} &\fs{\bf{0.016}}/\bl{0.55}\\
    & room2 & \bl{1.26}/\bl{20.2} & \bl{0.901}/\bl{8.71} & \fs{\bf{0.022}}/\fs{\bf{0.92}}\\
    & office0 & \bl{1.14}/\bl{20.1} & \bl{0.021}/\bl{0.71} & \fs{\bf{0.006}}/\fs{\bf{0.69}}\\
    & office1 & \bl{0.81}/\bl{36.3} & \fs{\bf{0.016}}/\bl{0.75} &\bl{0.017}/\fs{\bf{0.64}}\\
    & office2 & \bl{0.83}/\bl{19.9} & \bl{0.012}/\fs{\bf{0.40}} &\fs{\bf{0.007}}/\bl{0.44}\\
    & office3 & \bl{0.76}/\bl{18.8} & \bl{0.015}/\bl{0.67} & \fs{\bf{0.006}}/\fs{\bf{0.30}}\\
    & office4 & \bl{0.89}/\bl{46.3} & \bl{0.033}/\bl{0.82} & \fs{\bf{0.009}}/\fs{\bf{0.23}}\\
    \bottomrule
  \end{tabular}
  \vspace{-2.0em}
\end{table}

\begin{figure*}[!t]
  \centering
  \includegraphics[width=0.9\linewidth, trim={0 5cm 0 4.0cm}, clip]{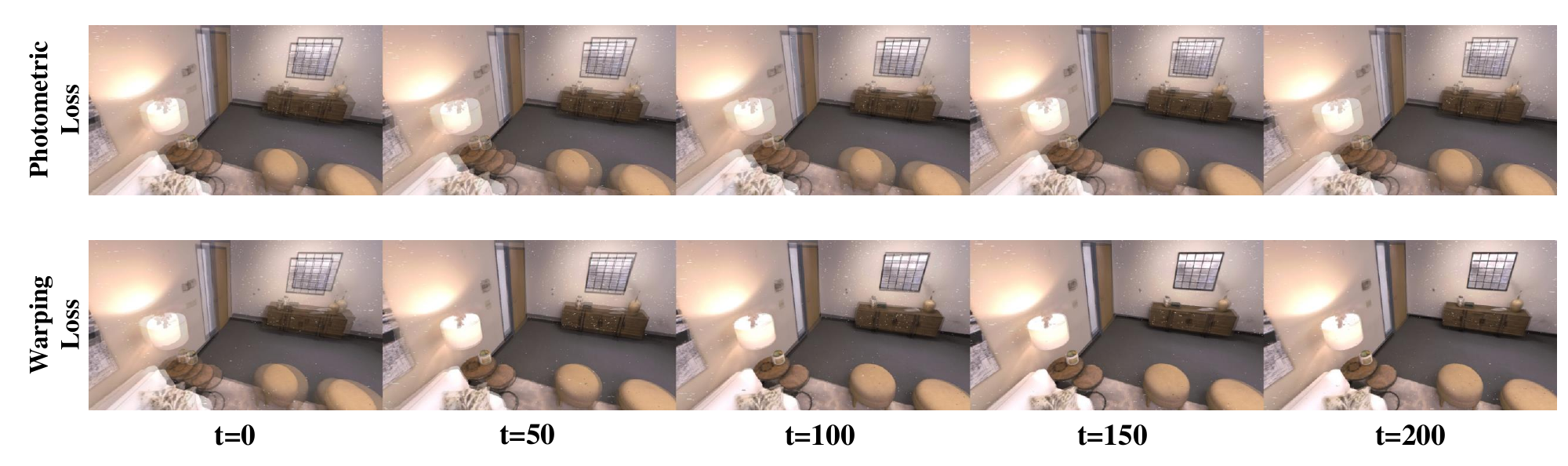}
  \vspace{-1.5em}
  \caption{\textbf{Pose estimation results.} We visualize the rendering images based on the estimated pose at time t and the query image to compare different optimization methods. 
  }
  \label{fig:optimization_step}
  \vspace{-2.0em}
\end{figure*}

We first compare our method with the SOTA structure-based method PixLoc~\cite{sarlin2021back} and the regression-based method CoordiNet~\cite{moreau2022coordinet} on the Replica dataset.  
Our method and PixLoc adopt 200 images as reference images,
while for CoordiNet, we use 2000 images for training to generate reasonable results. The evaluation results are shown in Table~\ref{SOTA table}. Our method achieves state-of-the-art results on the Replica dataset. We believe that PNeRFLoc surpasses PixLoc on the Replica dataset for two primary reasons: i) Each query image in Replica has a relatively large view-point change compared to the reference image, leading to a large initial reprojection error in PixLoc and causing the optimization to fall into incorrect local minima;
ii) with high-quality input images and accurate camera poses, PNeRFLoc can learn a fine NeRF model, which further facilitates rendering-based optimization given the initial pose estimation from the structure-based localization. 
The regression-based method CoordiNet has the worst performance unsurprisingly due to its poor generalization to the query image with large view-point changes although more reference images are provided to train the regression model. Please refer to our supplementary material for more comparisons.

\vspace{-0.5em}
\subsection{Evaluation on the Cambridge and 7Scenes}
\label{ssec:exp_renderbased}
\begin{table*}[h]
  \caption{\textbf{Comparison on the Cambridge Landmarks and 7Scenes datasets.} We report median translation/rotation errors(meters/degrees). Best results are highlighted as \fs{\bf{first}},~\nd{\bf{second.}}}
  \label{table:render}
  \centering
  \vspace{-0.8em}
  {\small
  \begin{tabularx}{0.92\linewidth}{lcccccccc}
    \toprule
    \multicolumn{1}{c}{} & Methods & PoseNet & CoordiNet & {$\quad$LENS} & DFNet & DSAC* & PixLoc & \textbf{PNeRFLoc}\\
    \midrule
    \multirow{8}{*}{\rotatebox{90}{7Scenes}} & Chess & \bl{0.32}/\bl{8.12} & \bl{0.14}/\bl{6.7} & \bl{0.03}/\bl{1.3} & \bl{0.04}/\bl{1.48} & \fs{\bf{0.02}}/\bl{1.10} & \fs{\bf{0.02}}/\fs{\bf{0.80}} & \fs{\bf{0.02}}/\fs{\bf{0.80}} \\
    & Fire & \bl{0.47}/\bl{14.4} & \bl{0.27}/\bl{11.6} & \bl{0.10}/\bl{3.7} & \bl{0.04}/\bl{2.16} & \fs{\bf{0.02}}/\bl{1.24}  & \fs{\bf{0.02}}/\fs{\bf{0.73}} & \fs{\bf{0.02}}/\nd{\bf{0.88}}\\
    & Heads & \bl{0.29}/\bl{12.0} & \bl{0.13}/\bl{13.6} & \bl{0.07}/\bl{5.8} & \bl{0.03}/\bl{1.82} & \fs{\bf{0.01}}/\bl{1.82}  & \fs{\bf{0.01}}/\fs{\bf{0.82}}& \fs{\bf{0.01}}/\nd{\bf{0.83}}\\
    & Office & \bl{0.48}/\bl{7.68} & \bl{0.21}/\bl{8.6} & \bl{0.07}/\bl{1.9} & \bl{0.07}/\bl{2.01} & \nd{\bf{0.03}}/\bl{1.15}  & \fs{\bf{0.03}}/\fs{\bf{0.82}} & \nd{\bf{0.03}}/\nd{\bf{1.05}}\\
    & Pumpkin & \bl{0.47}/\bl{8.42} & \bl{0.25}/\bl{7.2} & \bl{0.08}/\bl{2.2} & \bl{0.09}/\bl{2.26} & \nd{\bf{0.04}}/\nd{\bf{1.34}}  & \fs{\bf{0.04}}/\fs{\bf{1.21}} & \bl{0.06}/\bl{1.51}\\
    & Kitchen & \bl{0.59}/\bl{8.64} & \bl{0.26}/\bl{7.5} & \bl{0.09}/\bl{2.2} & \bl{0.09}/\bl{2.42} & \nd{\bf{0.04}}/\bl{1.68}  & \fs{\bf{0.03}}/\fs{\bf{1.20}} & \bl{0.05}/\nd{\bf{1.54}}\\
    & Stairs & \bl{0.47}/\bl{13.8} & \bl{0.28}/\bl{12.9} & \bl{0.14}/\bl{3.6} & \bl{0.14}/\bl{3.31} & \fs{\bf{0.03}}/\fs{\bf{1.16}}  & \nd{\bf{0.05}}/\nd{\bf{1.30}} & \bl{0.32}/\bl{5.73}\\
    \midrule
    \multirow{5}{*}{\rotatebox{90}{Cambridge}} & Kings & \bl{1.66}/\bl{4.86} & \bl{0.70}/\bl{2.92} & \bl{0.33}/\bl{0.5} & \bl{0.43}/\bl{0.87} & \nd{\bf{0.15}}/\bl{0.3}  & \fs{\bf{0.14}}/\fs{\bf{0.24}}& \bl{0.24}/\nd{\bf{0.29}}\\
    & Hospital & \bl{2.62}/\bl{4.90} & \bl{0.97}/\bl{2.08} & \bl{0.44}/\bl{0.9} & \bl{0.46}/\bl{0.87} & \nd{\bf{0.21}}/\bl{0.4}  & \fs{\bf{0.16}}/\fs{\bf{0.32}}& \bl{0.28}/\nd{\bf{0.37}}\\
    & Shop & \bl{1.41}/\bl{7.18} & \bl{0.73}/\bl{4.69} & \bl{0.27}/\bl{1.6} & \bl{0.16}/\bl{0.59} & \fs{\bf{0.05}}/\bl{0.3} & \fs{\bf{0.05}}/\fs{\bf{0.23}} & \nd{\bf{0.06}}/\nd{\bf{0.27}}\\
    & Church & \bl{2.45}/\bl{7.96} & \bl{1.32}/\bl{3.56} & \bl{0.53}/\bl{1.6} & \bl{0.50}/\bl{1.49} & \nd{\bf{0.13}}/\bl{0.4}  & \fs{\bf{0.10}}/\fs{\bf{0.34}}& \bl{0.40}/\bl{0.55}\\
    & Court & \bl{2.45}/\bl{3.98} & \bl{\textcolor{white}{0.00}}- & \bl{\textcolor{white}{0.00}}- & - & \nd{\bf{0.49}}/\bl{0.3}  & \fs{\bf{0.30}}/\fs{\bf{0.14}}& \bl{0.81}/\nd{\bf{0.25}} \\
    \bottomrule
  \end{tabularx}
  }
\end{table*}

\begin{figure}[t]
  \centering
  \vspace{0.6em}
  \includegraphics[width=0.4\textwidth, trim={0 0em 0 0.2cm}, clip]{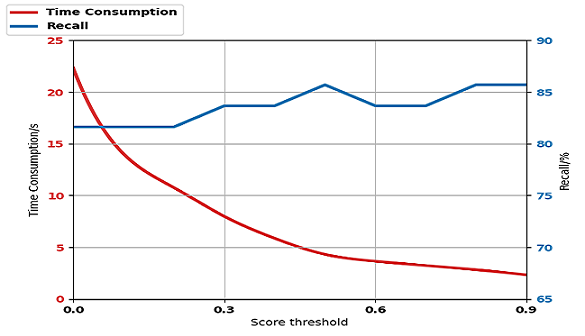}
  \vspace{-1em}
  \caption{We show the effectiveness of our reliable and repeatable score filtering on localization efficiency and accuracy.}
  \label{fig:score}
  \vspace{-1.9em}
\end{figure}

We compare with multiple SOTA approaches \cite{sarlin2021back, moreau2022coordinet, moreau2022lens, dsac++} on the benchmark visual localization datasets, \ie, Cambridge Landmarks and 7Scenes. We report the median translation 
and rotation error for each scene in Table~\ref{table:render}.

For the indoor 7Scenes dataset, since the generated depth by DSAC*~\cite{dsac++} are noisy with misalignments, it affects the training of the methods based on depth images, like DSAC* and our method. As a result, our method performs on-par or slightly worse than the SOTA PixLoc method on 7Scenes, while it still outperforms all other methods in general.  In the subsequent ablation experiments, we provide the results with relatively better depth inputs, which confirms our observations. 

For the outdoor Cambridge Landmarks dataset, there are large appearance variations and dynamic objects. 
Moreover, we find that the provided camera poses and intrinsic parameters of the training images are not very accurate, which prevents PNeRFLoc from learning a fine 3D NeRF model and rendering higher-quality novel view images. Even so, PNeRFLoc still performs on-par with the SOTA method and performs much better than all other NeRF-boosted localization methods \cite{moreau2022lens, chen2022dfnet}, which demonstrates its robustness and the potential of the NeRF-based methods for outdoor datasets. As long as more accurate depths and camera poses are provided, our method achieves the SOTA performance as 
shown on the Replica dataset. 

\vspace{-0.4em}
\subsection{Ablation Studies}

\begin{table*}[h]
  \vspace{-0.8em}
  \caption{\textbf{Ablation study.} We report the median translation/rotation errors (meters/degrees) and time consumption (seconds/per image). The best results are highlighted in \textbf{blod}.}
  \label{tab: ablation}
  \centering
  \vspace{-0.9em}
  \begin{tabular}{lcc}
    \toprule
    \multicolumn{1}{c}{\multirow{2}{*}{Config.}} & \multicolumn{2}{c}{Replica}\\
    \cmidrule(lr){2-3}
    \multicolumn{1}{c}{} & room0 & office0\\
    \midrule
    w/o Rendering-Based Optimization & 0.030 / 0.79 / \textbf{3.20} & 0.082 / 1.38 / \textbf{1.88}\\
    w/o Blank Depth Mask & 0.027 / 0.63 / 5.84 & 0.050 / 1.09 / 5.66\\
    w/o Warping Loss, w/ Photometric Loss & 0.035 / 0.81 / 47.7 & 0.082 / 1.38 / 39.2 \\
    Full Model & \textbf{0.005} / \textbf{0.29} / 5.56 & \textbf{0.006} / \textbf{0.56} / 5.45\\
    \bottomrule
  \end{tabular}
  \vspace{-0.2em}
\end{table*}

\label{ssec:exp_ablation}
\begin{table*}[!t]
  \vspace{-0.5em}
  \caption{\textbf{Comparison of using input depth and estimated depth.} We report median translation/rotation errors (meters/degrees) and the best results is highlighted as \fs{\bf{first}}.}
  \vspace{-0.8em}
  \label{depth table}
  \centering
  \begin{tabularx}{0.90 \textwidth}{lXXXXXX}
    \toprule
    \multicolumn{1}{c}{\multirow{2}{*}{Config.}} & \multicolumn{3}{c}{7Scenes} & \multicolumn{3}{c}{Replica}\\
    \cmidrule(lr){2-4} \cmidrule(lr){5-7}
    \multicolumn{1}{c}{} & chess & office & stairs & room1 & office0 & office1\\
    \midrule
    With Estimated Depth & \fs{\bf{0.02}}/\fs{\bf{0.80}} & \fs{\bf{0.03}}/\fs{\bf{1.05}} & \bl{0.32}/\bl{5.73} & \bl{0.02}/\bl{0.55} & \fs{\bf{0.01}}/\bl{0.56} & \bl{0.02}/\bl{0.64}\\
    With GT Depth & \fs{\bf{0.02}}/\fs{\bf{0.80}} & \fs{\bf{0.03}}/\bl{1.06} & \fs{\bf{0.20}}/\fs{\bf{3.61}} &
    \fs{\bf{0.01}}/\fs{\bf{0.34}} & \fs{\bf{0.01}}/\fs{\bf{0.54}} & \fs{\bf{0.01}}/\fs{\bf{0.39}}\\
    \bottomrule
  \end{tabularx}
  \vspace{-1.5em}
\end{table*}

\boldparagraph{Justification of the proposed rendering-based optimization.}
As shown in Fig.~\ref{fig:optimization_step} and Table~\ref{tab: ablation}, we justify our design decisions by comparing different variants of PNeRFLoc. All experiments were optimized 250 times during the rendering-based localization stage. We report the median translation/rotation errors (meters/degrees) and time consumption (seconds/per image). 
We can see that the proposed rendering-based optimization significantly improves the localization accuracy given the initial pose estimated in the structure-based estimation stage. 
Without the blank depth mask, the accuracy slightly degrades due to the numerous sampling points on the image, and a small proportion of blank area sampling does not affect the overall trend of optimization. Furthermore, direct optimization using photometric loss is more time-consuming, and it may also fall into incorrect local minima due to the backpropagation through networks. 
These ablation studies demonstrate the efficacy and efficiency of the proposed rendering-based optimization.

\begin{figure}[t]
  \centering
  \vspace{-0.4em}
  
  \includegraphics[width=0.4\textwidth, trim={0 0.3cm 0 0}, clip]{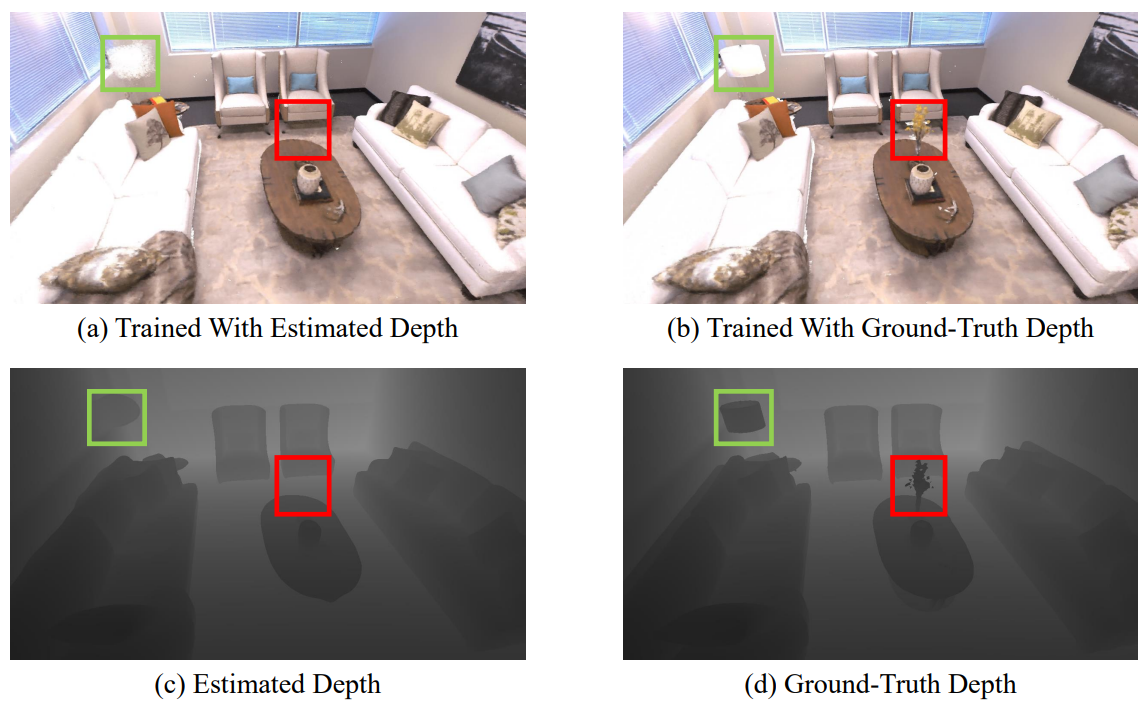}
  \vspace{-1.0em}
  \caption{We inspect the trained model with estimated depth or with input GT depth.}
  \label{fig:estimate depth}
  \vspace{-2.0em}
\end{figure}

\boldparagraph{Impact of score filtering.}
We analyzed the impact of score filtering on the Replica's office4 scene. As shown in Fig.\ref{fig:score}, we report the time consumption (in seconds) and recall at (5cm, 5\degree). With the increase of score threshold, the efficiency of the PnP algorithm is significantly improved, which is because the number of remaining candidate points in the point cloud decreases, saving the time to calculate cosine similarity. At the same time, we can see that the accuracy is preserved and even slightly improved with the score filtering because more reliable points for the scene are selected. 

\vspace{-0.2em}
\boldparagraph{Performance with input depth images. }
Since our method requires depth images to establish a point-based NeRF representation of the scene, we analyze the robustness of PixelLoc against the depth images.  
So as shown in Table~\ref{depth table}, we compare our results of learning the NeRF model with ground-truth input depth and estimated depth on the 7Scenes and Replica datasets. Since the Replica dataset is a synthetic dataset, its GT depth is dense and accurate, allowing for more precise neural point clouds and better rendering quality. In contrast, the estimated depth often loses details.
As illustrated in Fig.\ref{fig:estimate depth}, the estimated depth rendered by MonoSDF~\cite{yu2022monosdf} fails to capture the vase. Therefore, using GT depth on the Replica dataset significantly improves localization accuracy. In real-world scenes, however, the depth obtained by the Kinect RGB-D sensor is noisy, which is mainly influenced by the reflection and refraction of object surfaces, as well as the sensor's maximum and minimum measurement ranges. Consequently, 
the improvement in scene training quality is limited, and there is no significant improvement in localization accuracy. 
However, the advantage of taking the input depth is evident in the stairs scene, where the complex spatial structure leads to misalignment in the estimated depth. Please refer to our supp. material for more ablation studies.


\vspace{-0.5em}
\section{Conclusion}
In this paper, we present a novel visual localization method based on point-based neural scene representation. With a novel feature adaption module that bridges the features for localization and neural rending, the proposed PNeRFLoc enables 2D-3D feature matching for initial pose estimation and rendering-based optimization for pose refinement. Moreover, we also develop several techniques for efficient rendering-based optimization and robustness against illumination changes and dynamic objects. Experiments show the superiority of the proposed method by integrating both structure-based and rendering-based optimization, especially on the synthetic data suitable for NeRF modeling. 
Although our current framework is more efficient than the existing neural rendering-based optimization, we should further improve the efficiency and integrate it into visual odometry for real-time applications. 

\clearpage
\section{Acknowledgments}
This work was partially supported by the NSFC (No.~62102356).
We are also very grateful for the illustrations crafted by Lin Zeng.

\bibliography{aaai24}

\end{document}